\documentclass[10pt,twocolumn,letterpaper]{article}

\usepackage{wacv}
\usepackage{times}
\usepackage{epsfig}
\usepackage{graphicx}        
\usepackage{amsmath}
\usepackage{amssymb}
\usepackage{color}
\usepackage{tabularx}
\usepackage{booktabs}
\usepackage{multirow}
\usepackage{algorithm}
\usepackage{algorithmicx}
\usepackage{algpseudocode}  
\usepackage{array}
\usepackage{xspace}
\usepackage{balance}
\usepackage{enumitem}
\usepackage{mathtools}
\usepackage{rotating}
\usepackage{url}
\usepackage{tabulary}
\usepackage[table]{xcolor}
\usepackage{subcaption}
\usepackage[export]{adjustbox}
\usepackage{ctable}
\usepackage{diagbox}
\usepackage{subcaption}
\newcolumntype{Y}{>{\centering\arraybackslash}X}
\usepackage{textcomp}
\usepackage{authblk}
        
\def\Vec#1{{\boldsymbol{#1}}}

\DeclareMathOperator*{\argmin}{arg\,min}

\graphicspath{{./figs/}}

\usepackage[pagebackref=true,breaklinks=true,letterpaper=true,colorlinks,bookmarks=false]{hyperref}

\wacvfinalcopy 

\def\wacvPaperID{102} 

\ifwacvfinal\fi
\begin{document}

\title
	{
	Weakly-Supervised Multi-Person Action Recognition in 360$^{\circ}$ Videos\vspace{-1.9em}
	}
\author[1]{Junnan~Li\thanks{This work is mainly completed during Junnan Li's period of internship at NEC Corporation}}
\author[2]{Jianquan~Liu}
\author[1]{Yongkang~Wang}
\author[2]{Shoji~Nishimura}
\author[1]{Mohan~S.~Kankanhalli\vspace{-0.6em}}
\affil[1]{School of Computing, National University of Singapore}
\affil[2]{NEC Corporation}
\affil[ ]{\tt\small lijunnan@u.nus.edu, j-liu@ct.jp.nec.com, yongkang.wong@nus.edu.sg, s-nishimura@bk.jp.nec.com, mohan@comp.nus.edu.sg\vspace{-0.6em}}

\maketitle
\ifwacvfinal\thispagestyle{empty}\fi

\begin{abstract}
The recent development of commodity 360$^{\circ}$ cameras have enabled a single video to capture an entire scene,
which endows promising potentials in surveillance scenarios.
However, research in omnidirectional video analysis has lagged behind the hardware advances.
In this work,
we address the important problem of action recognition in top-view 360$^{\circ}$ videos.
Due to the wide filed-of-view,
360$^{\circ}$ videos usually capture multiple people performing actions at the same time.
Furthermore, the appearance of people are deformed.
The proposed framework first transforms top-view omnidirectional videos into panoramic videos using a calibration free method.
Then spatial-temporal features are extracted using region-based 3D CNNs for action recognition.
We propose a weakly-supervised method based on multi-instance multi-label learning,
which trains the model to recognize and localize multiple actions in a video using only video-level action labels as supervision.
We perform experiments to quantitatively validate the efficacy of the proposed method over state-of-the-art baselines and variants of our model,
and qualitatively demonstrate action localization results.
To enable research in this direction,
we introduce the 360Action dataset.
It is the first omnidirectional video dataset for multi-person action recognition with a diverse set of scenes, actors and actions.
The dataset is available at {\small \url{https://github.com/ryukenzen/360action}}.

\end{abstract}	

\section{Introduction}
\label{sec:introduction}

Omnidirectional cameras can monitor a vast scene with a small budget.
Recently, commodity omnidirectional cameras such as Samsung Gear 360 and Kodak PixPro SP360 have been developed, which can capture high-quality 4K videos.
A single top-view omnidirectional camera covers the same area as multiple conventional cameras,
making it a preferable device in surveillance scenarios.
Besides being cost-efficient and easier to install,
an omnidirectional camera requires only one algorithm to analyze the entire scene,
which avoids the inconvenience of synchronization and coordination among multiple conventional cameras,
and reduces security risks of privacy attack.

Despite the huge potential of omnidirectional cameras for video surveillance,
360$^{\circ}$ video analysis has received limited attention.
In this paper, we address the important problem of action recognition in 360$^{\circ}$ videos.
There are two challenges arising from 360$^{\circ}$ videos that make state-of-the-art deep network based action recognition algorithms ineffective.
First, the appearance of people are deformed. 
Specifically, people would be rotated at varying angles, thus making a deep network pretrained on standard perspective videos unable to extract useful features.
In this paper, we propose a method to transform an omnidirectional video into a panoramic video where people stand upright.
Our method is calibration-free, easy to implement, and does not require any training.

The wide field-of-view (FoV) of omnidirectional cameras results in the second challenge for action recognition.
In a practical scenario where the camera is installed at a place with large pedestrian volume,
the videos are likely to capture many people performing actions at the same time.
Since it is computational intensive to analyze each person individually, 
an efficient method should be able to simultaneously recognize actions for multiple people.
Furthermore, from the perspective of curating training data,
it is both expensive and time-consuming to extensively annotate each person's position (\ie~bounding box) and action.
On the other hand, it is much easier to acquire annotation only for the video-level action labels without linking each action to a specific person.

In this work,
we propose a weakly-supervised method for multi-person action recognition in high-resolution videos.
Our model is weakly-supervised in the sense that it is trained using only video-level action labels.
We formulate the problem as multi-instance multi-label (MIML) learning,
and exploit two intuitions to facilitate the design of our framework:
(1) \textit{only a fraction of regions in the video are informative for a certain action}, and
(2) \textit{one person can only perform one action at a time}.

Our key contributions can be summarized as follows:
\begin{itemize}[leftmargin=*]
	\setlength\itemsep{0pt}
	\item 
	We propose a novel framework for multi-person action recognition in top-view omnidirectional videos.
	The first step of our framework addresses people's rotational deformation by using a calibration-free method to transform omnidirectional videos into panaromic videos.
	\item
	The second step of our framework achieves multi-person action recognition with only video-level labels as supervision.
	We propose region-based 3D CNN to extract informative spatial-temporal features,
	and divide the spatial-temporal features into multiple instances for multi-label learning. 
	Our weakly-supervised model can learn to not only recognize but also localize each action in the video.
	\item 
	We introduce 360Action, the first omnidirectional video dataset for action recognition.
	Our dataset contains high-resolution videos recorded in diverse scenes.
	360Action paves the way for future research in omnidirectional video analysis.
	\item 
	We perform multi-person action recognition experiments on 360Action dataset.
	We quantitatively validate the efficacy of the proposed method,
	and qualitatively demonstrate action localization results.
	Furthermore, we conduct ablation study to analyze several model design choices. 
\end{itemize}

\section{Related Work}
\label{sec:literature}
\begin{figure*}[!t]
 \centering
  	\includegraphics[width=1\textwidth]{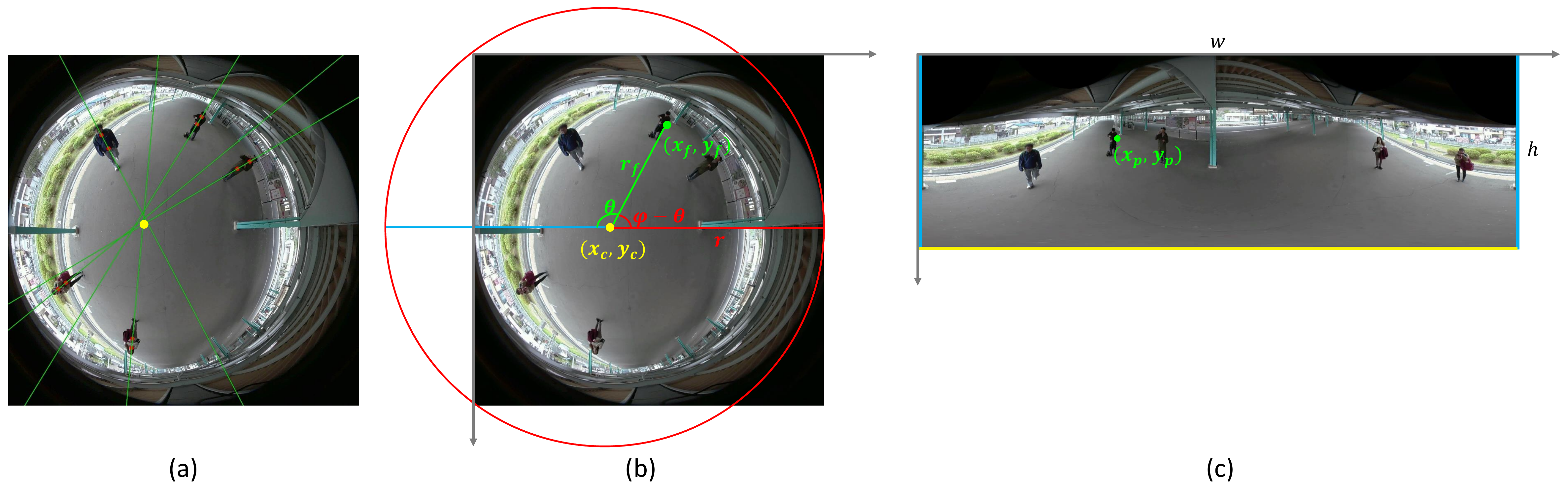}
  	\vspace{-4ex}
  \caption
  {
  \small
  (a) We locate the center (yellow) as the point closest to all spine lines (lines that are approximately perpendicular to the ground).
  (b\&c) We establish a mapping from any pixel $(x_p,y_p)$ in the panoramic frame to a pixel $(x_f,y_f)$ in the fisheye frame.   
  }
  \label{fig:transform}
 \end{figure*}  	

\subsection{Action Recognition} 
Action recognition is a long-standing problem in computer vision and has been extensively studied.
Recent approaches based on deep networks trained using large-scale video datasets have achieved great progress~\cite{Junnan_NIPS}.
One line of approaches use two-stream Convolutional Neural Networks (CNNs) to process RGB and optical flow frames as appearance and motion information, respectively~\cite{Simonyan_NIPS_2014,Wang_CVPR_2015,Christoph_NIPS_2016,Christoph_CVPR_2016,Feichtenhofer_CVPR_2017}.
Another line of approaches focus on CNNs with 3D convolutional kernels~\cite{Ji_PAMI_2013,Tran_ICCV_2015,Carreira_CVPR_2017}.
3D CNNs can effectively extract spatial-temporal features directly from videos.
Very recently, Hara~\etal~\cite{Hara_ICCV_2017,Hara_CVPR_2018} trained deep 3D CNNs using the large-scale Kinetics dataset~\cite{kay2017kinetics},
and achieve state-of-the-art action recognition performance on multiple benchmarks.
In this work,
we use 3D CNNs as our backbone architecture for spatial-temporal feature extraction.

\subsection{Omnidirectional Video Analysis} 
Due to the effectiveness of omnidirectional cameras for video surveillance,
researcher have studied omnidirectional pedestrian detection,
where the main challenge is pedestrian's rotational deformation.
Some works~\cite{Roman_arXiv_2018,Krams_AVSS_2017} propose to first transform omnidirectional images into perspective images and then apply standard pedestrian detectors.
However, the transformation relies heavily on calibrated camera parameters,
which requires user-interaction.  
Other works~\cite{Tamura_WACV_2019,Liu_arXiv_2017} train orientation-aware networks using pedestrian images with synthetic rotation.
However, such training introduces large computational cost and the trained network is biased towards the data used. 
In this paper, we propose a method to transform omnidirectional videos into panoramic videos where people stand upright.
Our method does not require camera calibration nor training.

Unlike conventional perspective videos, 
only a limited amount of work has studied action recognition in omnidirectional videos~\cite{Chen_ICMI_2002,Rainer_IJDET_2005,Ang_IJICIC_2012}.
Previous methods use constrained data and have limited performance.
Our work is the first to exploit the representation power of deep networks for multi-person action recognition in 360$^{\circ}$ videos.
We also introduce a new benchmark dataset to enable future research in this direction.

Besides the video domain, 
360$^{\circ}$ data has also been exploited to learn visual representations in a self-supervised manner~\cite{Junnan_MM} , 
which achieves improved performance on many vision tasks.
 

\subsection{Multiple Instance Learning} 
Multiple instance learning (MIL) is first introduced by Dietterich~\etal~\cite{MIl} for drug activity prediction.
In MIL framework,
each example consists of a bag of instances,
and only the bag-level label is available.
Recently, researchers have incorporated deep networks into the MIL framework for weakly-supervised image classification~\cite{Wu_MIL},
object detection~\cite{Cinbis_MIL,Hoffman_MIL},  co-saliency detection~\cite{Zhang_MIL}, social relationship recognition~\cite{Junnan_social},
and image segmentation~\cite{LSE}.

Multi-instance multi-label learning is a variant of MIL where an example with multiple instances is also associated with multiple labels~\cite{Zhou_MIML,Zhou_MIML2}.
In our problem,
we have multiple regions as instances from a video,
and only the video-level action labels are available as weak supervision.

\section{Omnidirectional to Panoramic}
\label{sec:transform}

To address the rotational deformation of people in omnidirectional videos,
we introduce a method to transform a top-view omnidirectional video into a panoramic one,
so that CNNs can be used for feature extraction.
Our method is \textit{calibration free}.
It does not require access to camera parameters or the configuration of the camera,
therefore is more applicable in practical scenarios.
Next we delineate the details.

First,
the dimensions of the panorama have a proportional relationship with the FoVs of the camera:
\begin{equation}
	\frac{h}{w}=\frac{\text{VFoV}}{2\times\text{HFoV}},
\end{equation}
where $w$ and $h$ are the width and height of the panorama,
respectively.
HFoV is the horizontal field-of-view (\ie,~360$^\circ$) of the fisheye camera
and VFoV is the vertical field-of-view (\ie,~235$^\circ$).
In this work we set $h=800$, which leads to $w=2451$ according to the above equation.

Our method exploits this observation:
in an omnidirectional frame captured by a top-view fisheye camera,
straight lines that are perpendicular to the ground would all intersect at a single point,
which we refer to as the center $c$.  
In order to locate its pixel coordinate $(x_c,y_c)$,
we detect the spines of people standing upright and use the spines to approximately represent lines perpendicular to the ground.
As shown in Figure~\ref{fig:transform}a,
we use Mask-RCNN~\cite{mask-rcnn} to detect two keypoints for each person: mid-shoulder and mid-hip.
Then we connect the two keypoints to acquire the spine for a person,
represented by the line $k_ix+y+z_i=0$.
Note that we rotate the frame by $\{0^{\circ},90^{\circ},180^{\circ},270^{\circ}\}$ and repeat the spine detection,
so that people at different orientations can be utilized.

Having detected $K$ spine lines,
we find $c$ as the point with the smallest total distance \textit{w.r.t} all lines:
\begin{equation}
	(x_c,y_c)=\argmin_{(x,y)}\sum_i^K \frac{|k_ix+y+z_i|}{\sqrt{k_i^{2}+1}}	
\end{equation} 
Note that we can compute $(x_c,y_c)$ for multiple frames and average the coordinates for a more accurate center location. 

Next, we need to find a mapping from a pixel $p=(x_p,y_p)$ in the panorama to its corresponding pixel $f=(x_f,y_f)$ in the fisheye frame.
In the fisheye frame (see Figure~\ref{fig:transform}b),
we denote the distance between $c$ and its furthest frame boundary as $r$, and the distance between $f$ and $c$ as $r_f$.
A user-defined angle ($\phi$) determines the blue starting 
line from which we unwrap the fisheye frame.
$\theta$ denotes the angle between the blue line and the green line.
We can establish a mapping from $(x_p,y_p)$ to $(\theta,r_f)$ as
\begin{equation}
\frac{x_p}{w}=\frac{\theta}{360}, \hspace{1ex}
\frac{h-y_p}{h}=\frac{r_f}{r}
\end{equation}
Then we can map the polar coordinate $(\theta,r_f)$ to the Cartesian coordinate $(x_f,y_f)$ with
\begin{equation}
\frac{x_f-x_c}{r_f}=\cos(\phi-\theta), \hspace{1ex}
\frac{y_c-y_f}{r_f}=\sin(\phi-\theta)
\end{equation}

Thus, we have successfully established the mapping $(x_p,y_p)\rightarrow(\theta,r_f)\rightarrow(x_f,y_f)$.
Note that if the corresponding $f$ for a $p$ is outside the fisheye frame,
we assign black color to $p$.
    
\section{Multi-person Action Recognition}
\label{sec:method}
\begin{figure*}[!t]
 \centering
  	\includegraphics[width=1\textwidth]{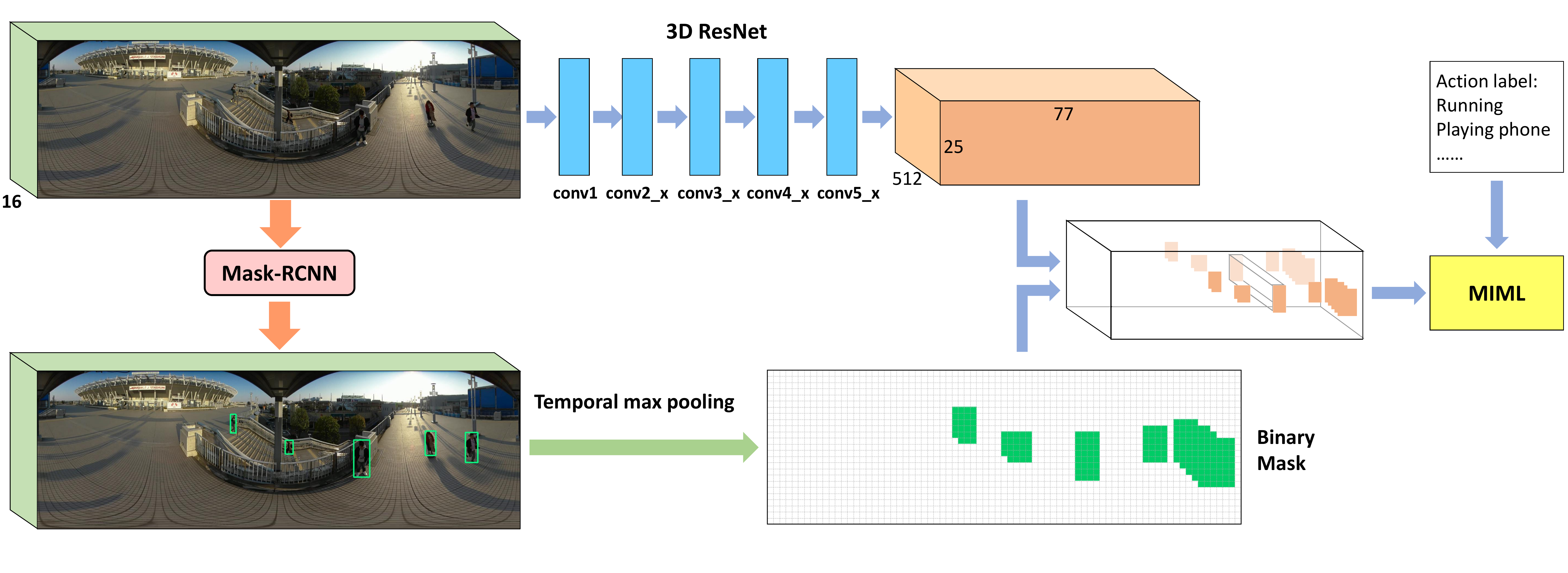}
  	\vspace{-4ex}
  \caption
  {
  \small
  Overview of the proposed action recognition framework. We use 3D ResNet to extract a convolutional feature map for a input video clip. Uninformative features are set as zeros by applying a binary mask calculated with person detection. The masked convolutional feature map is used for multi-instance multi-label learning.   
  }
  \label{fig:network}
 \end{figure*}  	
\begin{figure*}[!t]
 \centering
  	\includegraphics[width=1\textwidth]{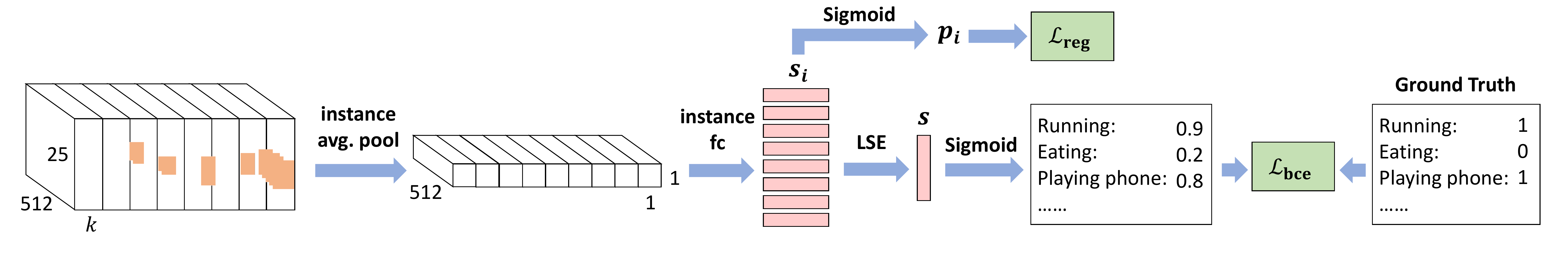}
  	\vspace{-4ex}
  \caption
  {
  \small
  The proposed Multi-instance Multi-label Learning (MIML) method. The masked convolutional feature map is divided into multiple instances. 
  Each instance outputs a set of scores $\Vec{s_i}$ for the possible actions.
  All instance-level scores are then aggregated with the Log-Sum-Exp (LSE) function.
  We add a regularization term ($\mathcal{L}_\mathrm{reg}$) to the cross entropy loss ($\mathcal{L}_\mathrm{bce}$), which penalizes the model if a single instance outputs high scores for multiple actions.
  }
  \label{fig:MIML}
 \end{figure*}  	
Given a high-resolution panoramic video containing multiple people,
we propose a method to recognize all actions in the video with one forward pass.
Our method is weakly-supervised,
where only video-level action labels are available during training.
An overview of the proposed framework is shown in Figure~\ref{fig:network}.
It consists of two steps:
spatial-temporal feature extraction and  multi-instance multi-label learning.
Next we explain our method in details.

\subsection{Region-based 3D CNN}
\label{sec:region}
Deep 3D CNNs trained on large-scale video datasets can learn representative spatial-temporal features~\cite{Hara_CVPR_2018}.
Therefore,
we utilize 3D ResNet-34~\cite{Hara_ICCV_2017} pretrained on Kinetics~\cite{kay2017kinetics} to extract features for each 16-frame clip.
Previous methods on action recognition normally resize the videos to heights of 240 pixels.
However,
since our videos have a much wider FoV and higher resolution than conventional videos,
such resizing would make the people extremely small and the actions unrecognizable.
Therefore, we use the original video with size $800\times2451$ as input.
The resulting convolutional feature map from the conv5\_x layer has a spatial size of $25\times77$.

A significant portion of the video contains only background and does not provide useful information for the actions.
We would like to discard those uninformative and potentially misleading features.
To this end,
we apply a person detector (\ie, Mask-RCNN~\cite{mask-rcnn}) to acquire the bounding boxes for all people in each frame (examples are provided in the supplementary material).
Then we max-pool the bounding boxes across all 16 frames in a clip to get a binary mask.
This mask is robust to false negative detection in certain frames because the person can be detected in other frames within the same clip.
We resize the mask to the same width and height as the convolutional feature map,
and multiply the feature map with the mask so that only features in the masked region are preserved while others are set as zero.
The masked spatial-temporal features are used for learning actions.

\subsection{Multi-instance Multi-label Learning}
\label{sec:MIML}

As shown in Figure~\ref{fig:MIML},
given a masked convolutional feature map,
we split it into $N$ blocks where each block has width $k$ (features are zero-padded).
We refer to each block as an instance $i$,
and apply spatial average pooling on each instance to acquire a feature of size $512\times1\times1$.
Each instance-level feature is then flattened and processed by a fully-connected (fc) layer to generate a set of scores $\Vec{s_i}=\{s_i^a\}, \forall a\in\mathcal{C}$ for the set of action classes $\mathcal{C}$.

For a ground-truth action $a$,
an obvious way to aggregate its instance-level scores is to take the average across all instances:
\begin{equation}
\label{eqn:avg}
s^a=\frac{1}{N}\sum_{i}^{N}s_i^a
\end{equation}

However, this would assign the same weights to all instances,
even the ones that are not relevant to the action.
In fact, 
only one or a few instances are responsible for the occurrence of $a$.
Another aggregation function is the max function:
\begin{equation}
\label{eqn:max}
s^a=\max_i{s_i^a}
\end{equation}

\begin{figure*}[!t]
 \centering
  	\includegraphics[width=1\textwidth]{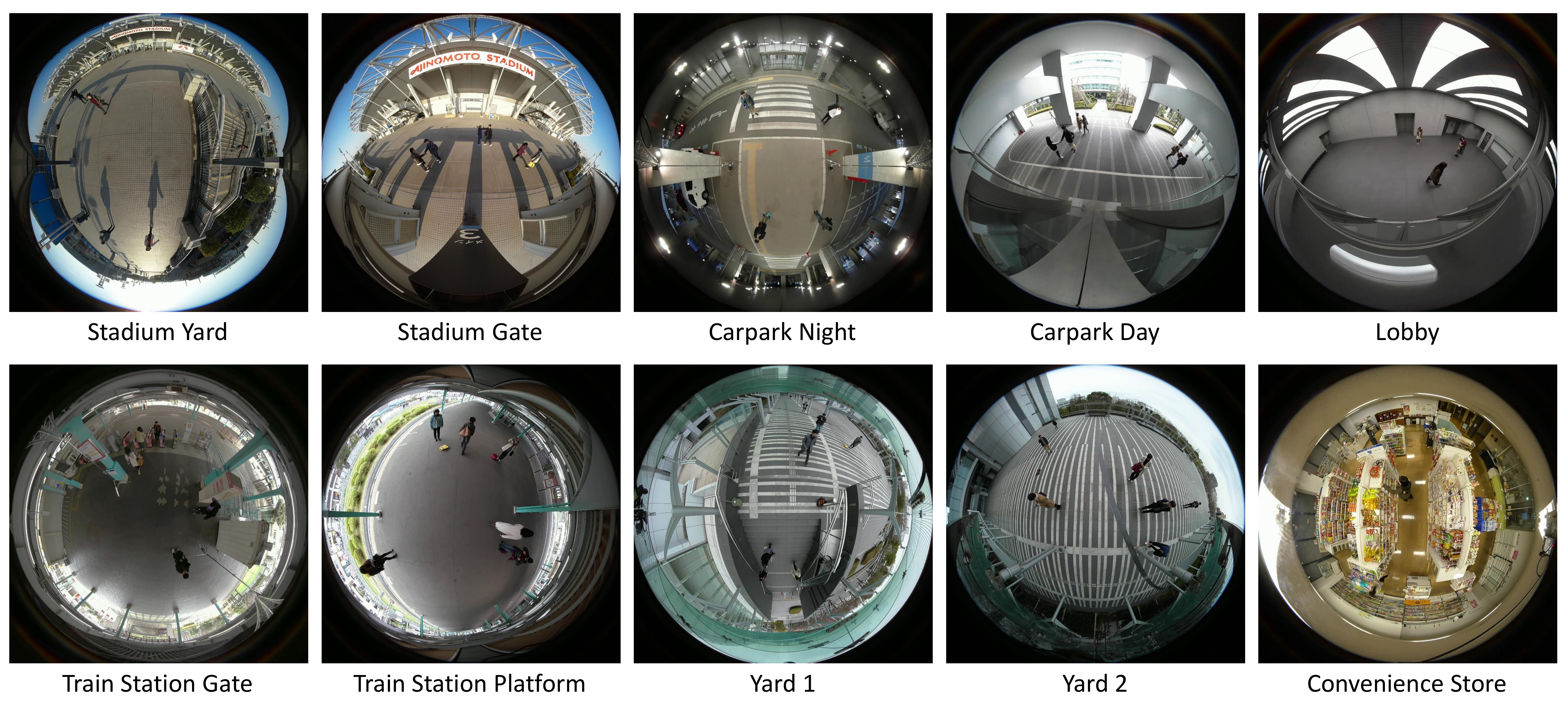}
  	\vspace{-4ex}
  \caption
  {
  \small
  Sample frames of the 10 diverse scenes in 360Action dataset. Each video captures multiple people performing actions.
  }
	\vspace{-1ex}
  \label{fig:dataset}
 \end{figure*}  	
This would consider only one instance to be responsible for the occurrence of the action,
neglecting the case where multiple people from different instances can perform the same action $a$.

To address this limitation, 
we use a smooth version of the max function,
called Log-Sum-Exp (LSE)~\cite{boyd2004convex}:
\begin{equation}
\label{eqn:lse}
s^a=\frac{1}{r}\log[\frac{1}{N}\sum_i^N \exp(rs_i^a)]
\end{equation}

We can use $r$ to control how smooth the function is.
Larger $r$ would have a similar affect to $\max$ and consider the most important instance,
whereas smaller $r$ would consider other less important instances. 
In our experiment we set $r=0.8$ (see Section~\ref{sec:experiment_action} for ablation study on hyper-parameters).

The aggregated scores are then scaled to $[0,1]$ with the sigmoid function:
\begin{equation}
p^a=\frac{1}{1+e^{-s^a}}
\end{equation}

During training,
the label $y_a$ for an action $a$ takes the value 1 if the action exists in the video and 0 otherwise.
We train our model to minimize the binary cross-entropy loss:
\begin{equation}
\mathcal{L}_\mathrm{bce}=-\sum_{a}\big(y^a\log(p^a)+(1-y^a)\log(1-p^a)\big)
\end{equation}

\noindent\textbf{Sparsity Regularization.}
Since one person can only perform one action at a time,
we propose a regularization term which penalizes the model if a single instance outputs high scores for multiple actions.
To this end, we first apply sigmoid function to scale each instance's score $s^a_i$ to $[0,1]$:
\begin{equation}
p^a_i=\frac{1}{1+e^{-s^a_i}}
\end{equation}

Then, the regularization term is defined as:
\begin{equation}
\mathcal{L}_\mathrm{reg} =\sum_i \frac{\sum_a p_i^a-\max_a p_i^a}{\max_a{p_i^a}}
\end{equation}

Minimizing $\mathcal{L}_\mathrm{reg}$ would encourage each instance to output a high score only for the action that it is most confident about.

During training, The total loss to minimize is:
\begin{equation}
\label{eqn:reg}
\mathcal{L}=\mathcal{L}_\mathrm{bce}+\alpha \mathcal{L}_\mathrm{reg},
\end{equation}
where $\alpha$ controls the strength of regularization and is set as 0.001 in our experiment (value determined by validation).

\subsection{Action Localization}
\label{sec:gradcam}
After training the 3D CNN following the proposed method,
we can localize the predicted actions by finding the areas in the convolutional feature map that are relevant to certain predictions.
We exploit Grad-CAM~\cite{gradcam} for weakly-supervised action localization,
which work as follows.

Given a test video clip,
the trained model outputs a set of action scores $\{p^a\}, \forall a\in\mathcal{C}$.
For each predicted action $a$ with $p^a>0.5$,
we apply back-propagation to calculate the gradient of $p^a$ with respect to the $k$-th feature map $A^k$ of the last convolutional layer (conv5\_x), \ie,~$\frac{\phi p^a}{\phi A^k}$.
The gradients are global-average-pooled to obtain a weight $\alpha_k^a$, which captures the importance of $A^k$ for action $a$:
\begin{equation}
\alpha_k^a=\frac{1}{Z}\sum_i\sum_j \frac{\phi p^a}{\phi A^k_{ij}}
\end{equation} 

The convolutional feature maps are then weighted combined to obtain a heatmap of the same size as $A^k$ (\ie,~$25\times 77$ for 3D ResNet-34):
\begin{equation}
H = ReLU(\sum_k \alpha_k^a A^k)
\end{equation}

High values at certain positions in the heatmap indicate positive influence of that area to the action of interest.
We resize $H$ to the same size as the panoramic video frames (\ie, $800\times2451$) to visualize the action localization results.
Examples are shown in Figure~\ref{fig:localize}.

\section{360Action Dataset}
\label{sec:dataset}

In this section, we introduce 360Action, 
the first omnidirectional video dataset for action recognition.
Different from conventional action recognition datasets,
360Action contains 360$^\circ$ videos which capture multiple person performing multiple actions at the same time.
The dataset contains 784 diverse videos,
recorded using a state-of-the-art omnidirectional camera,
Kodak PixPro SP360 4K,
which has a horizontal FoV of 360$^\circ$ and a vertical FoV of 235$^\circ$.
All videos have a high resolution of $2880\times2880$.
Next we delineate the details of our dataset.

\textbf{Scenes.} 
360Action consists of videos from 10 diverse scenes including 2 indoors scenes (lobby, convenience store) and 8 outdoor scenes (stadium yard, stadium gate,  carpark night, carpark day, train station gate, train station platform, yard 1 and yard 2).
Example frames from each scene are shown in Figure~\ref{fig:dataset}.
Different scenes have different environments (e.g.~train station platform has moving trains in the background whereas carpark has cars in the background) and different lighting conditions (e.g.~daytime sunny, daytime cloudy, indoor lighting, etc.).
The diversity of scenes puts high requirement on the algorithm to be robust to different conditions.

\textbf{Subjects.} 
We hired 80 different subjects for video recording.
40 subjects are male and 40 subjects are female.
The ages of the subjects range from 10 to 40 years old. 
Each subject is assigned a consistent ID number across all videos. 
We split subjects with IDs 1-60 into training set,
and subjects with IDs 61-80 into test set. 

\textbf{Actions.} 
The dataset contains 19 classes of daily actions,
including 15 single-person actions (eating, wearing jacket, walking, waving, etc.) and 4 interactions (pushing, handshaking, taking something, giving something).
During recording,
we assign each subject a scripted set of actions to perform.
Each subject performs each action at least once in a scene.
In each video,
multiple subjects perform multiple actions concurrently.
The maximum number of concurrent actions in a video is 7.
On average, each video contains 4 concurrent actions.

\section{Experiment}
\label{sec:experiment}

\subsection{Implementation Details}
Our model is trained using stochastic gradient descent (SGD) with a momentum of 0.9.
We use a batch size of 32 clips and a learning rate of 0.01.
The learning rate is decayed by half every 10 epochs,
and the model is trained for 50 epochs in total.
All hyper-parameters are determined via cross-validation.

\subsection{Action Recognition}
\label{sec:experiment_action}

\noindent\textbf{Comparison with state-of-the-art-methods.}
First, we compare the proposed model with multiple state-of-the-art action recognition methods~\cite{Bagautdinov_CVPR_2017,r-c3d,Hara_ICCV_2017,MiCT}.
We modify these methods for our task of weakly-supervised multi-person action recognition.
Specifically,
we apply multi-label classification loss for all baselines.
We also remove the detection loss in~\cite{Bagautdinov_CVPR_2017} because we do not assume to have detection ground-truth.
The comparison results are shown in Table~\ref{tbl:sota}.
We report the mean average precision (mAP) across all classes,
which is the standard evaluation metric that takes both precision and recall into account.
Our MIML-based method significantly outperforms the baselines due to its unique ability to discover associations between instances and actions.

\begin{table}[!t]
	\centering
	\caption
		{\small
			Comparison with state-of-the-art methods for weakly-supervised multi-person action recognition on 360Action dataset.
		}
		\label{tbl:sota}
	\begin{tabular}{l c} 
	\toprule
	Method & mAP (\%)\\
	\midrule
	Collective~\cite{Bagautdinov_CVPR_2017} & 61.27 \\
	3D ResNet~\cite{Hara_ICCV_2017} & 61.95  \\
	R-C3D~\cite{r-c3d} & 58.74 \\	
	MiCT~\cite{MiCT} & 62.18 \\	
	\midrule
	Ours & \textbf{70.12}\\
	\bottomrule
	\end{tabular}
	
\end{table}	
\newcommand\ang{80}
\begin{table*}[!t]
	\centering
	\caption
		{\small
			Per-class average precision of the 19 actions.
		}
		\label{tbl:AP}
	\resizebox{\textwidth}{!}{  
	\begin{tabularx}{\textwidth}{c Y Y Y Y Y Y Y Y Y Y Y Y Y Y Y Y Y Y Y} 	
	\toprule
	Method&
	\rotatebox{\ang}{Walk}&\rotatebox{\ang}{Run}&\rotatebox{\ang}{Wave Hand}&\rotatebox{\ang}{Pickup sth.}&\rotatebox{\ang}{Drop sth.}&\rotatebox{\ang}{Take off jacket}&\rotatebox{\ang}{Wear jacket}&\rotatebox{\ang}{Phone call}&\rotatebox{\ang}{Play phone}&\rotatebox{\ang}{Eat snack}&\rotatebox{\ang}{Walk upstairs}&\rotatebox{\ang}{Walk downstairs}&\rotatebox{\ang}{Drink water}&\rotatebox{\ang}{Tap out station}&\rotatebox{\ang}{Tap in station}&\rotatebox{\ang}{Handshake}&\rotatebox{\ang}{Take sth.}&\rotatebox{\ang}{Give sth.}&\rotatebox{\ang}{Push}\\
	\midrule
	Max-pool&87.6&77.2&\textbf{82.5}&55.6&40.6&59.0&53.8&\textbf{45.3}&\textbf{50.7}&\textbf{49.7}&74.2&59.9&48.4&\textbf{98.1}&88.0&62.8&85.4&56.0&69.6
	\\
	Ours& \textbf{89.8}&\textbf{93.8}&81.2&\textbf{65.4}&\textbf{51.4}&\textbf{69.4}&\textbf{67.0}&44.9&48.0&39.6&\textbf{77.9}&\textbf{79.6}&\textbf{48.5}&97.5&\textbf{95.4}&\textbf{63.7}&\textbf{86.3}&\textbf{56.7}&\textbf{73.5}
	\\
	\bottomrule
	\end{tabularx}
	}
\end{table*}	
\begin{table}[!t]
	\centering
	\caption
		{\small
		 Comparison with multiple variants of the proposed method on 360Action dataset. The results validate the efficacy of each proposed component (\ie,~region mask, MIML, sparsity regularization).
		}
		\label{tbl:action}
	\begin{tabular}{l c} 
	\toprule
	Method & mAP (\%)\\
	\midrule
	Without mask & 67.52\\
	Avg-pool& 64.73\\
	Max-pool& 65.50\\
	MIML-avg& 68.59\\
	MIML-max& 69.07\\
	MIML-attention& 69.46\\
	Without $\mathcal{L}_\mathrm{reg}$& 69.25\\
	\midrule
	Ours (MIML-LSE)& \textbf{70.12}\\
	\bottomrule
	\end{tabular}
	
\end{table}	

\noindent\textbf{Ablation study.}
Next,
we conduct ablation study by comparing the proposed model with its multiple variants.
For each baseline, we remove one component from the proposed model while keeping others components unchanged,
so that we can examine the effect of each proposed component for weakly-supervised multi-person action recognition.
The variants are delineated as follows.
\begin{itemize}[leftmargin=*]
	\setlength\itemsep{0pt}
	\item
	Without mask: we do not use the binary region mask (see Section~\ref{sec:region}) to filter out useless information. In this case the MIML module receives a noisy input.
	\item
	Avg-pool: we do not use the proposed MIML module (see Section~\ref{sec:MIML}) to acquire the action scores. 
	Instead,
	we apply average pooling on the masked feature map from conv5\_x layer,
	and use a fully-connected layer to transform the avg-pooled feature into action scores.
	\item
	Max-pool: we do not use the proposed MIML module to acquire the action scores.
	Instead,
	we apply max pooling on the masked feature map from conv5\_x layer,
	and use a fully-connected layer to transform the max-pooled feature into action scores.	
	\item 
	MIML-avg: we use the average function (see Eqn.~\ref{eqn:avg}) rather than LSE to aggregate instance-level action scores.
	\item 
	MIML-max: we use the max function (see Eqn.~\ref{eqn:max}) rather than LSE to aggregate instance-level action scores.	
	\item 
	MIML-attention: we use the attention pooling method~\cite{TC3D} to aggregate instance-level action scores.
	For each instance, a scalar attention weight is generated by feeding its feature to a fully-connected layer.
	The attention weights across all instances are than normalized using a Softmax function.
	\item
	Without $\mathcal{L}_\mathrm{reg}$: we remove the regularization term that forces sparsity in instance-level scores (see Eqn.~\ref{eqn:reg}) .
\end{itemize}

Table~\ref{tbl:action} shows the comparison results.
The proposed MIML framework achieves a significant improvement of $+4.52\%$ over the standard feature max-pooling method.
Comparing the four instance-level score aggregation methods,
LSE performs better than attention pooling, max and average. 
Using region mask improves the performance by $+2.5\%$.
The proposed sparsity regularization can further improve the performance by $+0.87\%$.

In Table~\ref{tbl:AP},
we show the per-class average precision (AP) for our model and the max-pool baseline.
Generally speaking,
actions that involves subtle movements (\eg,~play phone, eat snack) have lower APs,
whereas actions with larger motions (\eg,~run, wave hand) are more easily recognized by the model.

\begin{figure}[!t]
 \centering
  	\includegraphics[width=0.9\columnwidth]{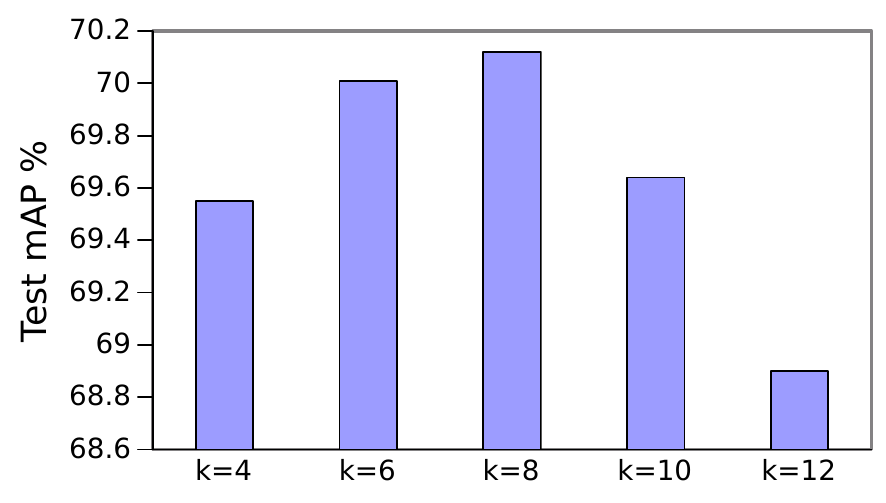}
  	\vspace{-2ex}
  \caption
  {
  \small
The effect of $k$ (\ie,~the width of each instance in the conv5\_x feature map) on action recognition performance.
  }
  \label{fig:k}
 \end{figure}  	
\begin{figure}[!t]
 \centering
  	\includegraphics[width=0.9\columnwidth]{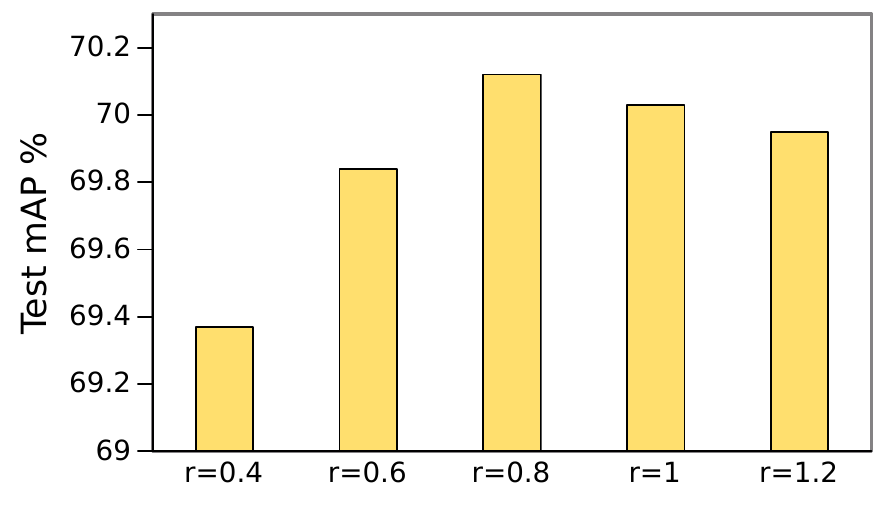}
  	\vspace{-2ex}
  \caption
  {
  \small
The effect of $r$ (which controls how smooth the LSE function is) on action recognition performance.
  }
\vspace{-1ex}
  \label{fig:r}
 \end{figure}  	

\begin{figure*}[!t]
 \centering
  	\includegraphics[width=1\textwidth]{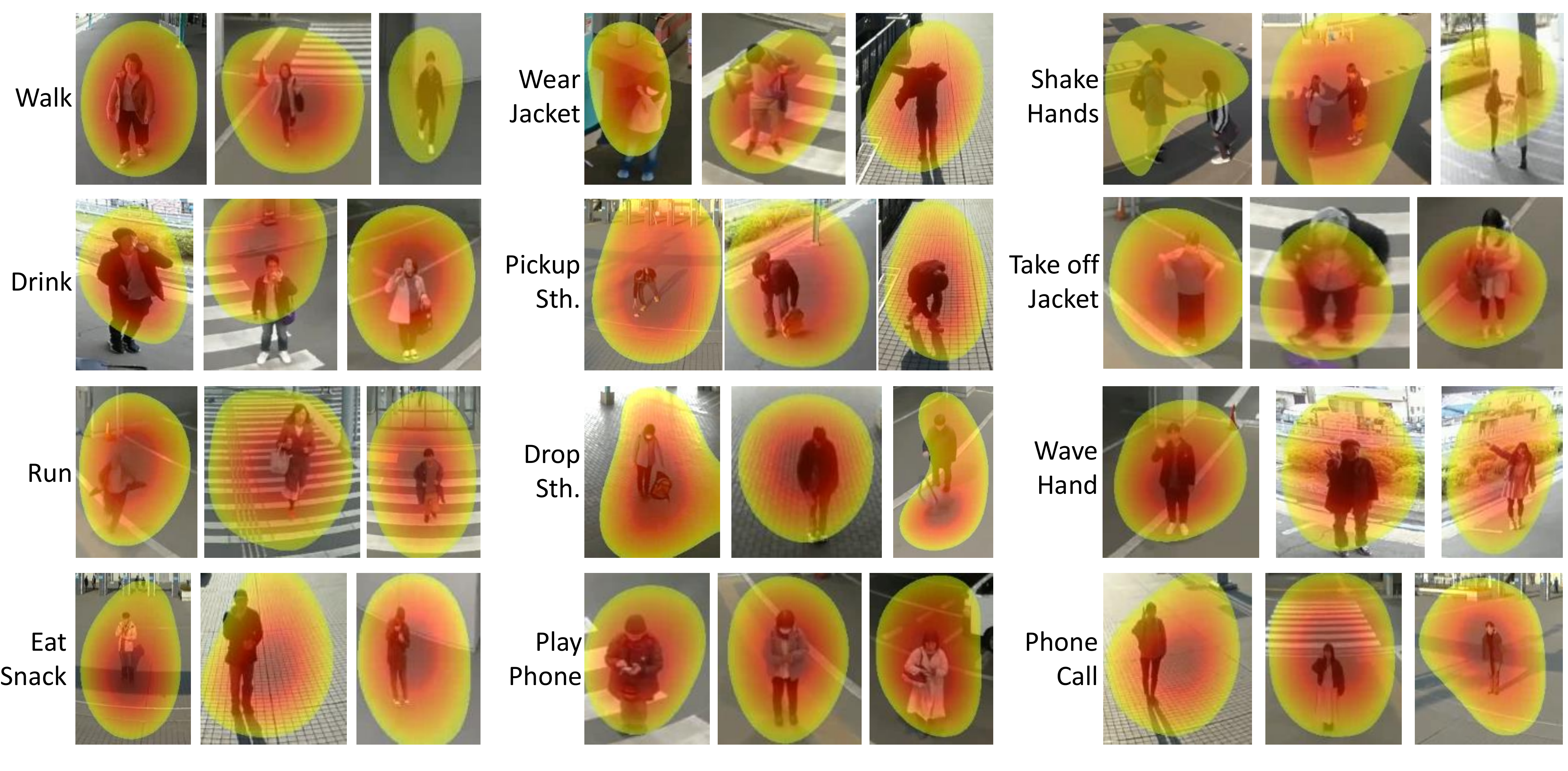}
  	\vspace{-4.5ex}
  \caption
  {
  \small
  Qualitative results for weakly-supervised action localization. We employ GradCAM~\cite{gradcam} to generate heatmaps that highlight the important areas for predicting certain actions. 
  Our model trained with only video-level labels learns to not only predict the multiple actions but also coarsely localize each of its predicted actions.
  }
\vspace{-1ex}
  \label{fig:localize}
 \end{figure*}

\noindent\textbf{Hyper-parameters.}
Here we examine the effects of two important hyper-parameters in our MIML framework:
$k$, the width of each instance in the conv5\_x feature map;
and $r$, the parameter that controls how smooth the LSE aggregation function is (see Eqn.~\ref{eqn:lse}). 

Figure~\ref{fig:k} shows the result of our method using different values of $k$.
Larger $k$ would result in fewer instances and more actions per instance,
whereas smaller $k$ leads to more instances with many instances containing no action.
$k=8$ achieves the best performance in our experiment.

Figure~\ref{fig:r} shows the result of our method using different values of $r$.
Larger $r$ would make the LSE function more similar to max whereas smaller $r$ would make it more similar to average.
In our experiment $r=0.8$ achieves the best performance.

\begin{table}[!t]
	\centering
	\caption
		{\small
			Inference speed analysis using different models for spatial-temporal feature extraction and person detection.
		}
		\label{tbl:speed}
	\resizebox{\columnwidth}{!}{  
	\begin{tabular}{l l | c c} 
	\toprule
	Feature Extractor & Person Detector & FPS & mAP (\%)\\
	\midrule
	3D ResNet-18 & YOLO & 6.28 & 66.17\\
	3D ResNet-34 & Mask-RCNN &1.14 & 70.12\\
	\bottomrule
	\end{tabular}
}

\end{table}

\noindent\textbf{Speed Analysis.}
During training,
we do not finetune the 3D CNN and only perform feature extraction once for the entire training set.
Therefore, 
the training process can be completed within 2 hours using a single NVIDIA V100 GPU.
However,
the inference speed is limited by the high resolution of the input videos.
We perform an inference speed analysis in Table~\ref{tbl:speed}.
By using a shallower feature extractor (\ie~3D ResNet-18) and a single-stage person detector (\ie~YOLO~\cite{yolo}),
we can achieve a much faster inference speed of 6.28 FPS at the cost of lower recognition performance.

\subsection{Action Localization}

We follow the method in Section~\ref{sec:gradcam} to calculated heatmaps for our model's predicted actions.
In Figure~\ref{fig:localize},
we show qualitative results of the heatmaps overlaid on cropped video frames.
Despite being trained using only action labels without any location information about where each action takes place,
the proposed model can learn to associate its predicted actions to specific people in the input videos,
thus achieving weakly-supervised action localization.

\section{Conclusion}
\label{sec:conclusion}

To conclude,
this paper aims to fill the gap between the rapid hardware development of $360^\circ$ cameras and the limited progress in omnidirectional video analysis.
We recognize the significant potential of $360^\circ$ cameras in surveillance scenarios, and address the important task of action recognition in top-view omnidirectional videos.
In face of the new challenges brought by the deformed and high-resolution videos,
we propose a framework that achieves multi-person action recognition in one forward pass.
To mitigate the difficulty of acquiring dense annotations,
our method learns to recognize and localize actions in a weakly supervised manner where only video-level labels are required in training.
In addition,
to facilitate research in this direction,
we curated 360Action,
the first $360^\circ$ video dataset with a diverse set of scenes and actors.
We hope to see future research in omnidirectional videos to achieve the same level of success as in conventional videos.
\section*{Acknowledgment}
\label{sec:acknowledgement}
This publication has emanated from the research at NEC Corporation, and is also partially supported by the National Research Foundation, Prime
Minister's Office, Singapore under its Strategic Capability Research
Centres Funding Initiative.

{\small
\balance
\bibliographystyle{ieee}
\bibliography{bib}
}

\end{document}